\documentclass[11pt]{article}

\usepackage[preprint]{acl}

\usepackage{times}
\usepackage{latexsym}

\usepackage[T1]{fontenc}

\usepackage[utf8]{inputenc}

\usepackage{microtype}

\usepackage{inconsolata}

\usepackage{graphicx}

\usepackage{multirow}
\usepackage{enumitem}
\usepackage{amsmath}
\usepackage{amssymb}
\usepackage{booktabs}

\usepackage{colortbl} 
\usepackage{hyperref}
\usepackage{subcaption}
\usepackage{ulem}
\usepackage{wrapfig}
\usepackage{fontawesome}
\usepackage{bbm}
\usepackage{pifont}
\newcommand{\tablestyle}[2]{\setlength{\tabcolsep}{#1}\renewcommand{\arraystretch}{#2}\centering\footnotesize}
\definecolor{LightCyan}{RGB}{232,241,255}
\definecolor{COLOR_MEAN}{HTML}{f0f0f0}

\title{TrainExplorer: Autonomous LLM Post-Training \\ via Plan-and-Decouple Exploration}

\author{Xuannan Liu$^{1,2}$ \quad
  Tian Lan$^{2}$ \quad
  Zekun Li$^{3}$ \quad
  Xing Cui$^{1}$ \quad \\
  \textbf{Zhengfeng Yu$^{4}$} \quad 
  \textbf{Feihu Jiang$^{2}$} \quad
  \textbf{Qianghuai Jia$^{2}$} \quad
  \textbf{Longyue Wang$^{2}$} \quad \\
  \textbf{Zhao Xu$^{2}$} \quad 
  \textbf{Weihua Luo$^{2}$} \quad
  \textbf{Peipei Li$^{1}$\thanks{Corresponding author.}} \quad
  \textbf{Ran He$^{3,5}$} \quad \\
  $^{1}$ Beijing University of Posts and Telecommunications \quad 
$^{2}$ Alibaba Group \quad \\
$^{3}$ Minzu University of China \quad 
$^{4}$  Beijing Institute of Technology \quad \\
$^{5}$ Center for Research on Intelligent Perception and Computing, NLPR, CASIA
  }

\begin{document}
\maketitle

\begin{abstract}
LLM-based agents can automate model post-training, accelerating progress toward recursive self-improvement. 
Despite their promise, existing approaches remain under-explored and face two main challenges: (1) \textbf{strategy lock-in} that limits exploration beyond the initial strategy; and (2) \textbf{coupled execution} that obscures the effects of individual modifications.
To address these challanges, we propose \textbf{TrainExplorer}, a multi-agent framework for autonomous post-training, that incorporates strategy planning and execution decoupling, to achieve broaden strategy exploration.
First, strategy planning draws on scientific literature to construct a candidate search space of training strategies.
Then, decoupled execution enables single-component interventions and attribution to guide strategy exploration.
Extensive experiments on PostTrainBench demonstrate that TrainExplorer achieves a weighted aggregate score of 39.02, outperforming Claude Code using the same LLM by 7.32 and surpassing GPT-5.6 Sol-powered Codex, which scores 36.23.
Code will be available.
\end{abstract}

\section{Introduction}
\label{sec:intro}
Recent advances in LLM-based agents have demonstrated strong capabilities in code generation~\cite{swebench,livecodebench}, tool use~\cite{schick2023toolformer,patil2025bfcl}, and long-horizon task execution~\cite{yao2023react, reflexion, lats}, accelerating the automation of AI research and development~\cite{wijk2025rebench,ward2025ctrlaltdeceit}.
Automated AI-driven researchers can already propose ideas, run experiments, evaluate outcomes, and derive critiques~\cite{lu2026aiscientist, starace2025paperbench}.
Such iterative loop can also support the development of more capable and efficient successor models, enabling recursive self-improvement~\cite{chan2026intelligenceexplosion}.
This progress underscore the importance of autonomous model training.

\begin{figure}[t!]
    \centering
    \includegraphics[width=\linewidth]{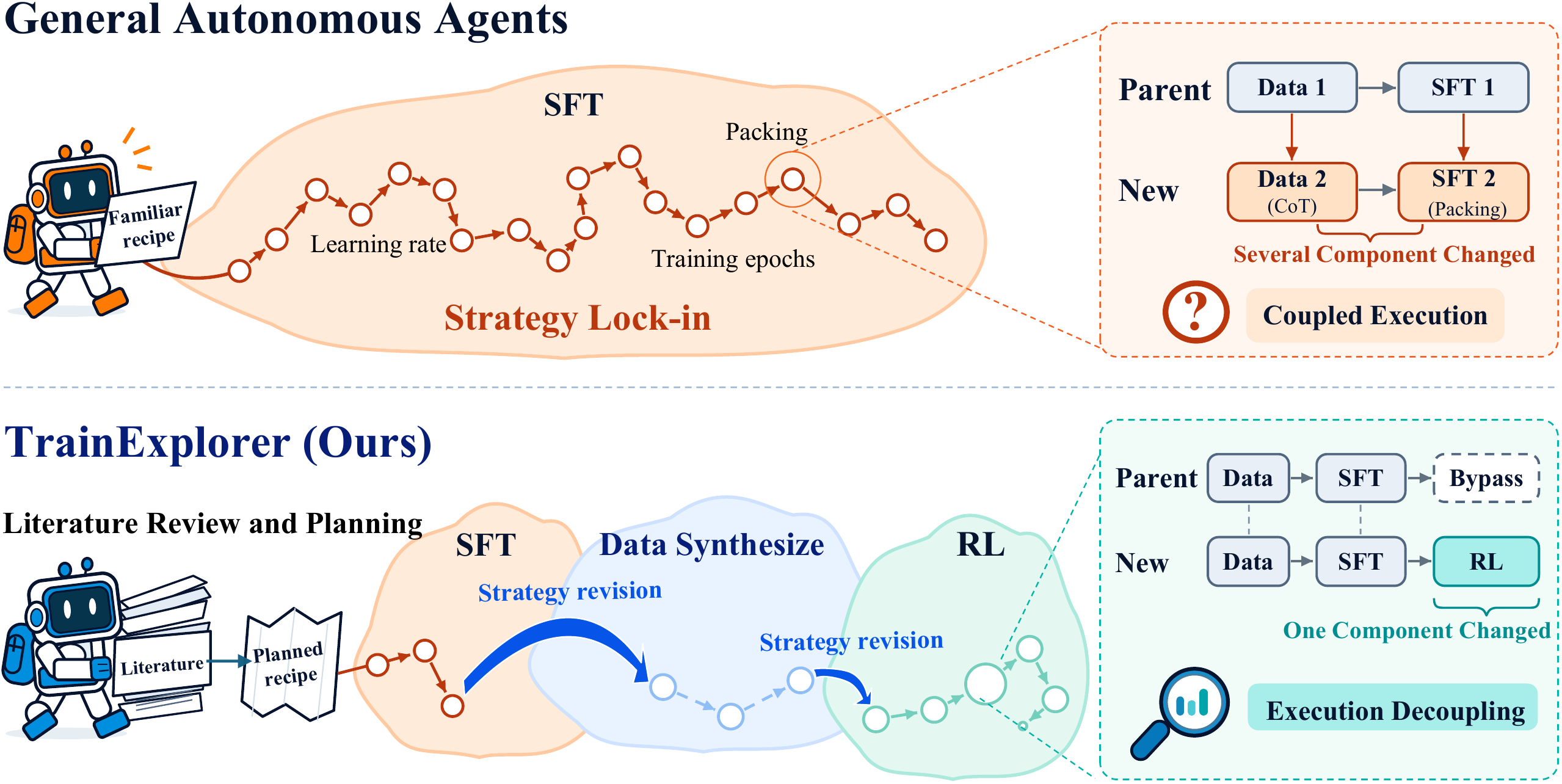}
    \caption{Comparison between a general autonomous agent and TrainExplorer. \textbf{Top:} The agent exhibits strategy lock-in by restricting exploration to refinements of its initial SFT strategy, and coupled execution by introducing CoT data examples and enabling training packing in the same experiment. \textbf{Bottom:} TrainExplorer uses strategy planning to construct a candidate strategy space and execution decoupling to enable single-component interventions and effect attribution.}    
    \label{fig:intro}
\end{figure}

Research on automated model training primarily focuses on machine learning (ML) engineering and LLM optimization.
Several studies have employed agents to automate complete ML pipelines~\cite{trirat2025automlagent,fang2025mlzero,chan2025mlebench,chen2026mars}.
In contrast, autonomous LLM training is more open-ended and complex, remaining largely underexplored.
Within LLM optimization, data-centric approaches focus on data selection~\cite{liu2024selectit} and data synthesis~\cite{zhou2024staragents}, while model-centric approaches investigate hyperparameter search~\cite{fang2026llmzero} and algorithm discovery~\cite{lu2024discopop}.
Recent works have also integrated data preparation, training, and eval evaluation into end-to-end workflows~\cite{rank2026posttrainbench,yano2025lamdagent, ma2026trex,yu2026autotrainess}.

Despite effective on target tasks, existing methods still faces two key challenges:
(1) \textbf{Strategy Lock-in}:
When strategy selection relies primarily on the LLM's internal knowledge, agents tend to favor familiar, widely used methods.
Thus, subsequent iterations often remain confined to local adjustments within the initial strategy, leaving other viable training approaches underexplored~\cite{lim2026missing,du2026rsimaster}.
As shown in Figure~\ref{fig:intro} \textbf{Top}, the agent's experiments remain within the initial supervised fine-tuning (SFT) strategy without exploring other training strategies.
(2) \textbf{Coupled Execution}:
Execution-level changes within a single experiment often simultaneously modify data-processing pipelines and training configurations, making it difficult to attribute the observed performance change to individual modifications~\cite{nam2025mlestar,chen2026mars}..
As shown in Figure~\ref{fig:intro} \textbf{Top}, the agent adds chain-of-thought (CoT) data examples and enables training-time packing in the same experiment, obscuring the individual effects of these two changes.

To address these challenges, we introduce \textbf{TrainExplorer}, a multi-agent framework for autonomous training that integrates strategy planning and execution decoupling to accelerate exploration toward an optimal training scheme.
TrainExplorer orchestrates specialized agents to propose strategies, execute experiments, and attribution analysis in an iterative training loop.
As shown in Figure~\ref{fig:intro} \textbf{Bottom}, to broaden strategy exploration, TrainExplorer builds a structured training experience libraries through literature search, from which a planning agent retrieves and selects applicable method to construct a candidate strategy space.
Furthermore, decoupled execution separates data processing and model training into independently editable components, enabling single-component interventions for controlled comparisons and effect attribution for distilling reusable experience.

We evaluate our framework on PostTrainBench~\cite{rank2026posttrainbench}, a benchmark designed for autonomous LLM post-training.
Our experiments demonstrate the superiority of the proposed TrainExplorer, which achieves a weighted aggregate score of 39.02, outperforming Claude Code using the same agent LLM by 7.32 and surpassing Codex powered by GPT-5.6 Sol which scores 36.23.
Moreover, TrainExplorer achieves a strategy switch rate of 37.5\%, substantially higher than Claude Code's 9.5\%, indicating that it can explore training schemes beyond its initial strategy. Our main contributions are as follows:
\begin{itemize}
    \item We introduce TrainExplorer, a \textbf{Plan-and-Decouple} framework for autonomous LLM post-training to to guide exploration toward more effective training strategies.
    \item We propose strategy planning to construct a candidate strategy space, and execution decoupling to enable single-component intervention and effect attribution.
    \item Extensive experiments on PostTrainBench demonstrate the effectiveness of TrainExplorer and show that the proposed Plan-then-Decouple design enables broader exploration of training strategies.
\end{itemize}

\section{Related Work}
\label{sec:realted_work}

\paragraph{Automated Model Training}
Research on automated model training spans traditional ML engineering and LLMs. 
For ML tasks, existing works operate under fixed-data assumptions, with agents primarily performing feature engineering, model search, and hyperparameter tuning~\cite{guo2024dsagent,trirat2025automlagent,nam2025mlestar,fang2025mlzero,liu2026mlagent,chen2026mars,huang2024mlagentbench,qiang2025mledojo,wijk2025rebench}.
For LLMs, automation methods primarily address both data-centric and model-centric optimization.
Data-centric methods modify data distributions by selecting task-relevant subsets, synthesizing supervision signals, and coordinating complete data-processing pipelines~\cite{xia2024less,wang2023selfinstruct,chen2024datajuicer,du2026datamaster,zhao2026andes}. 
Model-centric methods adjust training strategies through architecture design, hyperparameter search, and algorithmic discovery~\cite{cheng2025genesys,lu2024discopop}.

More recently, automated pipelines curate data, execute training, and evaluate outcomes across entire training loops~\cite{yano2025lamdagent,yu2026autotrainess}. 
Notably, a standardized testbed for LLM post-training, PostTrainBench, is provided to requiring agents to submit fully trained models within strict time budgets~\cite{rank2026posttrainbench}.
Recent work on PostTrainBench provides structured training interfaces~\cite{yu2026autotrainess}, reusable data skills~\cite{zhao2026andes}, and systematic experiment management~\cite{du2026rsimaster}.
While promising, strategies across these pipeline stages are often tightly coupled, making it challenging to isolate the specific causal impact of individual changes.
To bridge this gap, TrainExplorer decomposes strategy implementations into independently editable components, enabling rigorous effect tracing and guide subsequent refinement.

\paragraph{Recursive Self-Improvement}
Recursive self-improvement iteratively improves systems through proposal generation, experimental validation, and feedback-driven learning~\cite{chan2026intelligenceexplosion}.
Prior work applies this process to scientific discovery~\cite{lu2026aiscientist}, harness design~\cite{zhang2026dgm}, and model training~\cite{rank2026posttrainbench}.
Recent systems guide strategy refinement by comparing candidate implementations and analyzing their execution feedback~\cite{lee2026metaharness}.
However, reliance on a model's internal knowledge may limit subsequent iterations to local adjustments of the initial strategy, leaving viable alternatives underexplored~\cite{lim2026missing,du2026rsimaster}.
Thus, TrainExplorer uses scientific literature to construct an explicit strategy space, enabling systematic comparison and switching among training strategies throughout experiments.

\section{Problem Formulation}
Given a base model, a target task, a fixed evaluator, and a resource budget, autonomous LLM post-training aims to find a solution that maximizes performance on the target task through iterative experiments:
\begin{equation}
    s^{*}=\arg\max_s\mathcal{E}(\theta_s).
\end{equation}
Here, $s$ denotes a training solution specifying data acquisition and model training operations.
The model obtained by executing $s$ is denoted by $\theta_s$, and $\mathcal{E}(\theta_s)$ is its evaluation score under the fixed evaluator $\mathcal{E}$.
Under the resource budget, the agent can execute and evaluate only a finite number of candidate solutions and returns the model with the highest score among the evaluated candidates.

\begin{figure*}[t!]
    \centering
    \includegraphics[width=\linewidth]{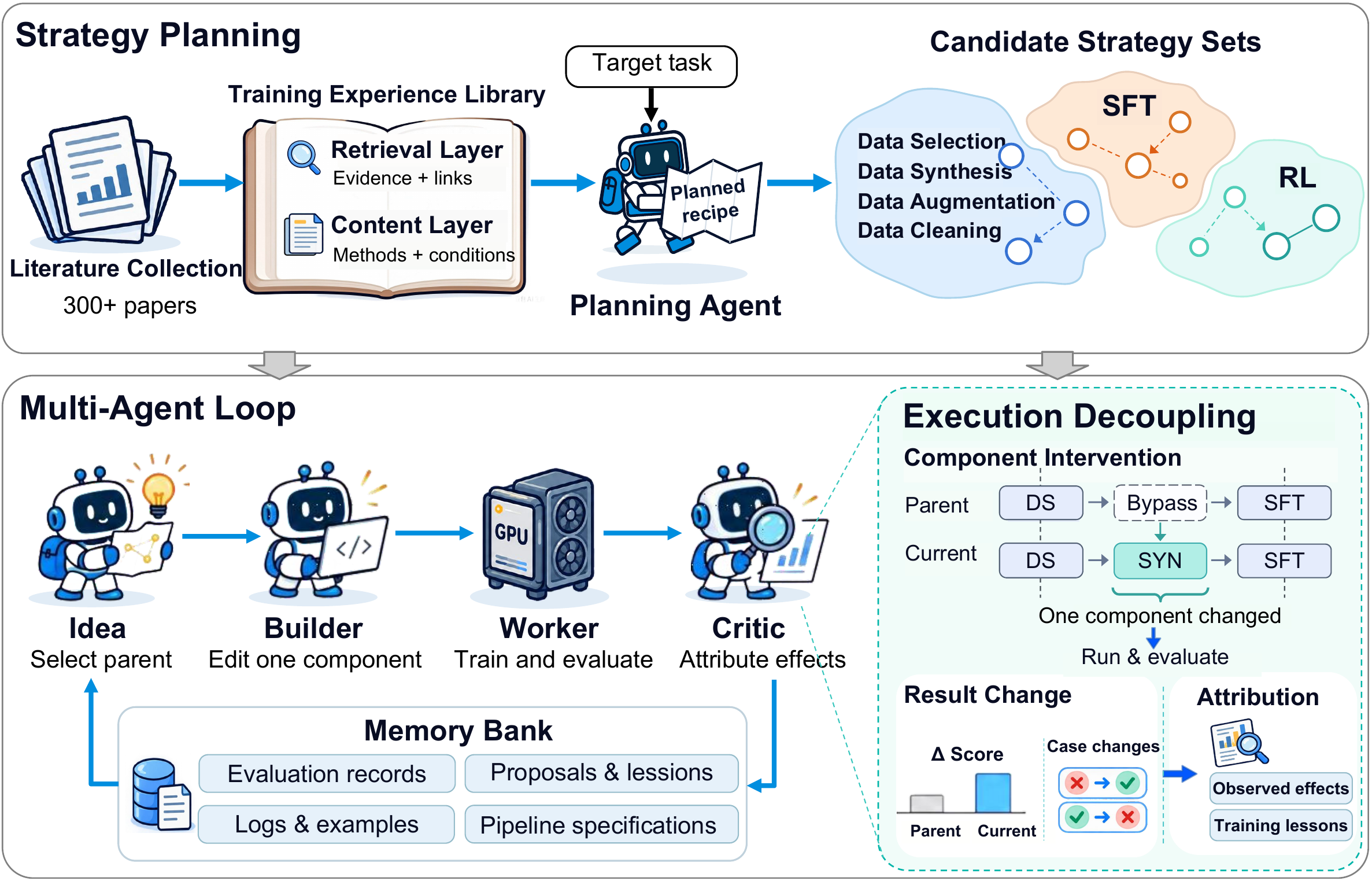}
    \caption{Overview of TrainExplorer, a multi-agent framework for autonomous LLM post-training that integrates strategy planning and execution decoupling. First, strategy planning draws on a structured knowledge base of scientific literature to construct a candidate strategy space. Then, the Idea, Builder, and Critic form an iterative loop for ideation, experimentation, and critique. Within this loop, execution decoupling enables single-component interventions and attributes their effects through controlled comparisons to guide subsequent exploration.}
    \label{fig:method}
\end{figure*}

\section{TrainExplorer}
\label{sec:method}
In this section, we present a multi-agent framework named TrainExplorer which incorporates strategy planning and execution decoupling for autonomous LLM post-training. 
We first introduce the overall framework. 
Then, we elaborate on our proposed strategy planning for building candidate search space.
Finally, we detail the execution decoupling for single-component intervention and effect attribution.

\subsection{Overall Framework}
\label{sec:method-overview}
As illustrated in Figure~\ref{fig:method}, TrainExplorer comprises a initial strategy planning stage followed by a multi-agent recursive loop.
During strategy planning, task-relevant data-processing and training strategies are selected from a structured library of prior training experiences to formulate the candidate strategy space $\mathcal{S}$.
TrainExplorer then searches this space for a optimal training solution:
\begin{equation}
    s^{*}=\arg\max_{s\in\mathcal{S}}\mathcal{E}(\theta_s).
    \label{eq:task-objective}
\end{equation}
To navigate this space, the multi-agent loop iteratively performs strategy proposal, experiment execution, and empirical analysis.
Within this loop, execution decoupling separates data construction and model training into independent editable components. 
Single-component interventions and effect attribution distill experimental feedback into reusable insights to guide strategy exploration.

\paragraph{Multi-Agent Loop.} Three specialized agents collaboratively drive this recursive loop:
\begin{itemize}
    \item \textbf{Idea Agent} selects a parent iteration and proposes a strategy intervention, conditioning the target task, the candidate strategy space, and attribution summaries from prior experiments.
    \item \textbf{Builder Agent} instantiates the intervention in the selected parent iteration as executable data-processing and training code. It then dispatches data preparation, model training, and outcome evaluation to a \textbf{Worker} process.
    \item \textbf{Critic Agent} assesses the intervention's efficacy by examining component-level modifications with shifts in evaluation scores and sample-level behaviors. It distills these observations into reusable training experiences.
\end{itemize}

\subsection{Strategy Planning}
\label{sec:method-strategy-space}
Strategy planning constructs a candidate strategy space $\mathcal{S}$ using a structured library of training experience $W$.
The library organizes empirical experience from the literature into searchable indices and granular entries, facilitating on-demand retrieval of relevant information.
Leveraging this library, the planning agent extracts candidate data-processing and training strategies tailored to the task.

\paragraph{Literature Collection.}
To equip TrainExplorer with prior post-training experience, we collect relevant literature to compile a experience library.
We curate literature based on dataset curation, and model training, considering both domain relevance and task-required capabilities. 
These tasks span seven domains: competitive, healthcare, and general mathematics, natural sciences, programming, tool, and writing. 
We screen candidate papers based on citation counts and venue prestige, yielding a collection of over 300 relevant papers.

\paragraph{Library Organization.}
To avoid context window exhaustion, we structure the collected training literature into a two-tiered library $W$, partitioned into a retrieval layer and a content layer~\cite{tang2026wikiskill}. 
This architecture enables progressive disclosure, loading knowledge strictly on demand. 
The retrieval layer maintains an index $I(W)$ of paper titles and concise evidence summaries. 
By encapsulating core dataset properties or methodological mechanisms, these summaries allow the agent to efficiently assess task relevance. 
Upon selection, the content layer supplies the detailed descriptions, experimental setups, and empirical results necessary for in-depth analysis.

\paragraph{Strategy Construction.}
The Planning Agent draws on the training experience library $W$ to construct a task-specific candidate strategy space of data construction and model training.
It first reads the knowledge index $I(W)$, identifies entries relevant to the task's domain and required capabilities, and uses tools to inspect detailed entries on demand.
For data construction, the agent examines knowledge coverage, task difficulty, and supervision; for model training, it assesses how each method's core mechanism aligns with the required capabilities.
It further evaluates whether each candidate strategy can be adapted to the current model and training setup within the resource budget.

Specifically, given the training experience library $W$, the Planning Agent selects $M$ strategy types based on the task requirements, such as data selection, data synthesis, data augmentation, data cleaning, SFT, or RL, and constructs a candidate set for each type:
\begin{equation}
    \left\{
        \left\{T_{\mathrm{strategy}}^{i,j}\right\}_{j=1}^{N_i}
    \right\}_{i=1}^{M}
    = A_{\mathrm{planning}}(W),
    \label{eq:strategy-retrieval}
\end{equation}
where $A_{\mathrm{planning}}$ denotes the Planning Agent, $M$ is the number of selected strategy types, $N_i$ is the number of candidates for the $i$-th type, and $T_{\mathrm{strategy}}^{i,j}$ describes its $j$-th candidate.
The Idea Agent selects component strategies from these candidate sets and combines them with specific configurations to construct or refine a post-training pipeline.

\subsection{Execution Decoupling}
\label{sec:method-component-attribution}
Execution decoupling organizes data processing and model training into independently editable components.
Each iteration modifies a single component while retaining the others, enabling the Critic Agent to relate changes in evaluation scores and response behavior to the intervention.
The resulting attribution summaries capture reusable training insights for subsequent iterations.

\paragraph{Component Intervention.}
TrainExplorer performs a single-component intervention at each iteration, modifying one component of the training pipeline while keeping the implementations of all other components fixed.
First, we decouple data processing and model training into independent components.
Each component encapsulates a method implementation and its configuration, with an input/output contract specifying data formats and fields to support independent modification and composition.
Second, validated component implementations are cached for reuse across iterations, reducing redundant code generation.

Specifically, iteration $t$ starts from a selected parent pipeline $s_k$ and adds, refines, or replaces a single component to obtain $s_t$.
All other component implementations and configurations are inherited from $s_k$.
We denote the resulting component change by $\delta_{k,t}$, which specifies the intervention analyzed in subsequent effect attribution.

\paragraph{Effect Attribution.}
TrainExplorer distills experimental feedback from single-component interventions into reusable training lessons.
By comparing the current pipeline with its parent, we relate changes in evaluation results and sample-level behavior to the component intervention, informing subsequent strategy exploration.

Specifically, after executing the training pipeline $s_t$ from iteration $t$, we compare the results of iterations $k$ and $t$ on the same evaluation set.
The change in evaluation score is
\begin{equation}
    \Delta s_{k,t}
    = \mathcal{E}(\theta_{s_t})-\mathcal{E}(\theta_{s_k}),
    \label{eq:performance-change}
\end{equation}
where $\theta_{s_t}$ and $\theta_{s_k}$ denote the models obtained from pipelines $s_t$ and $s_k$, respectively.
The corresponding per-example changes in sample-level behavior are denoted by $\Delta b_{k,t}$.
We combine these observations with the component intervention $\delta_{k,t}$ to generate an attribution summary:
\begin{equation}
    T_{\mathrm{attr}}^{t}
    = A_{\mathrm{critic}}\!\left(\delta_{k,t},\Delta s_{k,t},\Delta b_{k,t}\right),
    \label{eq:evidence-grounded-attribution}
\end{equation}
where $A_{\mathrm{critic}}$ denotes the Critic Agent.

The attribution summary $T_{\mathrm{attr}}^{t}$ captures the observed effects of the intervention and distills them into reusable training lessons.
We maintain a four-level memory bank that stores:
(1) evaluation records;
(2) strategy proposals and distilled lessons;
(3) execution logs and per-example evaluation results; and
(4) pipeline specifications.
Entries across these levels are linked by iteration, grounding the distilled lessons in the corresponding experimental evidence and implementation details.

\begin{table*}[t]
\tablestyle{5pt}{1.20}
\resizebox{\linewidth}{!}{%
\begin{tabular}{cc|ccccccc|c}
\toprule
\textbf{Method} & \textbf{Model} & AIME & ArenaHard & BFCL & GPQA & GSM8K & HealthBench & HumanEval & AVG \\
\midrule \midrule
\multicolumn{2}{c|}{Base Models} & 1.67 & 1.26 & 1.50 & 8.48 & 20.43 & 9.49 & 12.81 & 7.53 \\
\multicolumn{2}{c|}{Official Instruct Models} & 29.17 & 70.21 & 85.00 & 36.22 & 87.00 & 43.32 & 71.49 & 51.14 \\
\midrule
\rowcolor{COLOR_MEAN}
\multicolumn{10}{l}{\textit{General CLI Agent}} \\
Codex & GPT-5.6 Sol & 7.50 & 26.40 & 94.38 & 30.14 & 69.48 & 27.39 & 63.95 & 36.23 \\
Claude Code & Opus 4.8 & 10.83 & 36.09 & 47.13 & 28.35 & 68.87 & 31.68 & 47.33 & 32.90 \\
Claude Code & Kimi K3 & 11.67 & 36.71 & 9.67 & 25.23 & 76.32 & 31.84 & 62.30 & 31.96 \\
Claude Code & Opus 4.7 & 6.39 & 19.59 & 76.92 & 27.48 & 60.72 & 17.81 & 42.08 & 28.56 \\
Codex & GPT-5.5 & 3.75 & 12.89 & 56.13 & 30.67 & 56.70 & 19.79 & 48.86 & 27.23 \\
Cursor & Grok 4.5 & 10.42 & 14.13 & 1.50 & 27.07 & 65.12 & 11.32 & 51.30 & 23.45 \\
OpenCode & Gemini 3.1 Pro & 3.89 & 7.42 & 63.59 & 18.53 & 45.51 & 16.33 & 40.19 & 21.99 \\
Codex & GPT-5.4 & 0.56 & 9.57 & 23.59 & 27.98 & 48.18 & 13.87 & 27.34 & 19.00 \\
\midrule
\rowcolor{COLOR_MEAN}
\multicolumn{10}{l}{\textit{Autonomous PostTraining Agent}} \\
ANDES & GLM-4.7 & 3.90 & 25.66 & 85.67 & 27.12 & 72.04 & 31.51 & 48.74 & 33.39 \\
AutoTrainess & GPT-5.4 & 1.67 & 4.65 & 96.75 & 29.19 & 57.17 & 16.05 & 38.26 & 26.94 \\
\midrule
Claude Code & GLM-5.2 & 7.78 & 26.98 & 47.92 & 29.65 & 72.71 & 26.55 & 52.49 & 31.70 \\
Meta-Train & GLM-5.2 & 4.17 & 33.31 & 70.75 & 29.63 & 73.37 & 25.01 & 51.83 & 32.86 \\
\rowcolor{LightCyan}
\textbf{TrainExplorer} & GLM-5.2 & \textbf{11.39} & \textbf{34.74} & \textbf{92.00} & \textbf{31.79} & \textbf{78.48} & \textbf{30.89} & \textbf{59.04} & \textbf{39.02} \\
\bottomrule
\end{tabular}
}
\caption{Performance (\%) of post-training agents on PostTrainBench, averaged across four base models. Base and official instruct models are included as references. AVG denotes the weighted aggregate score across seven tasks. Bold indicates the best score in each column among methods using GLM-5.2 as the agent LLM.}
\label{tab:main-results}
\end{table*}

\begin{table*}[t]
\tablestyle{5pt}{1.20}
\resizebox{\linewidth}{!}{%
\begin{tabular}{cc|ccccccc|c}
\toprule
\shortstack{\textbf{Planning}} & \shortstack{\textbf{Decoupling}} & AIME & ArenaHard & BFCL & GPQA & GSM8K & HealthBench & HumanEval & AVG \\
\midrule \midrule
\ding{55} & \ding{51} & 1.11 & 23.80 & 93.00 & 28.72 & 73.89 & 20.22 & 57.32 & 32.51 \\
\ding{51} & \ding{55} & 3.33 & 17.24 & 91.00 & 29.99 & 75.92 & 25.04 & 62.80 & 34.22 \\
\rowcolor{LightCyan}
\ding{51} & \ding{51} & \textbf{6.67} & \textbf{27.66} & \textbf{94.00} & \textbf{32.07} & \textbf{81.05} & \textbf{27.36} & \textbf{65.24} & \textbf{37.77} \\
\bottomrule
\end{tabular}
}
\caption{Ablation of strategy planning (Planning) and execution decoupling (Decoupling) in TrainExplorer using Qwen3-1.7B-Base. \ding{51}/\ding{55} indicates whether each module is enabled. AVG denotes the weighted aggregate score across seven tasks. Bold indicates the best score in each column.}
\label{tab:ablation-results}
\end{table*}

\section{Experiments}
\label{sec:experi}

\subsection{Experimental Setup}

\paragraph{Datasets.}
We evaluate on PostTrainBench~\cite{rank2026posttrainbench}, which comprises seven benchmarks: AIME 2025, GSM8K~\cite{cobbe2021training}, HumanEval~\cite{chen2021evaluating}, BFCL v3~\cite{patil2025bfcl}, GPQA~\cite{rein2023gpqa}, ArenaHard v2~\cite{li2025arenahard}, and HealthBench~\cite{arora2025healthbench}. 
Each benchmark is paired with four base models: Qwen3-1.7B, Qwen3-4B~\cite{yang2025qwen3}, SmolLM3-3B~\cite{bakouch2025smollm3}, and Gemma3-4B~\cite{gemmateam2025gemma3}, yielding 28 configurations. 
We follow the official protocol, allocating a single 80 GB flagship GPU and a 10-hour time budget to each benchmark.

\paragraph{Metrics.}
Following PostTrainBench, we report both per-task scores and the weighted aggregate score. 
We first average each task's scores across the four base models, then aggregate across the seven tasks using the official task weights. 
These weights are computed by normalizing the inverse score gaps between the official instruct and base models for each task.


\paragraph{Baselines.}
We compare two categories of baselines:
(1) \textbf{General-Purpose CLI Agents}: Claude Code (Claude Opus 4.8, Claude Opus 4.7, Kimi K3, and GLM-5.2),
Codex CLI (GPT-5.6 Sol, GPT-5.5, and GPT-5.4),
Cursor (Grok 4.5), and OpenCode (Gemini 3.1 Pro).
(2) \textbf{Autonomous Post-Training Agents}: frameworks
tailored for optimizing post-training strategies, including
Meta-Train, ANDES, and AutoTrainess.
Meta-Train adapts Meta-Harness~\cite{lee2026metaharness} through a
propose--train--evaluate loop.
ANDES~\cite{zhao2026andes} provides an interactive
data synthesis tool.
AutoTrainess~\cite{yu2026autotrainess} provides structured
interfaces that incorporate human training expertise.

\subsection{Main Results}
To evaluate the effectiveness of TrainExplorer for autonomous LLM post-training, we compare it with general-purpose CLI agents and autonomous post-training agents on PostTrainBench.
Table~\ref{tab:main-results} presents the scores on seven tasks and the weighted aggregate score, averaged across the four base models.
Results for individual base models are provided in Appendix~\ref{sec:main-results-by-model}.
Our main findings are as follows:

\paragraph{Comparison with Base Models.}
TrainExplorer improves the weighted aggregate score of the base models from 7.53\% to 39.02\%, demonstrating substantial gains on the target tasks within the allotted budget.
On BFCL, TrainExplorer achieves 92.00\%, surpassing the 85.00\% achieved by the official instruction-tuned models.
Nevertheless, the overall score remains below the official instruction-tuned models' 51.14\%, highlighting substantial room for improvement in autonomous post-training.

\paragraph{Comparison with General-Purpose CLI Agents.}
TrainExplorer achieves a higher weighted aggregate score than all general-purpose CLI baselines, outperforming the strongest baseline, Codex (GPT-5.6 Sol), by 2.79 percentage points.
When both methods use GLM-5.2 as the agent LLM, TrainExplorer outperforms Claude Code by 7.32 percentage points.
This may be attributed to the Plan-and-Decouple design, which broadens strategy exploration and attributes effects to individual changes.

\begin{figure*}[!ht]
    \centering
    \includegraphics[width=\linewidth]{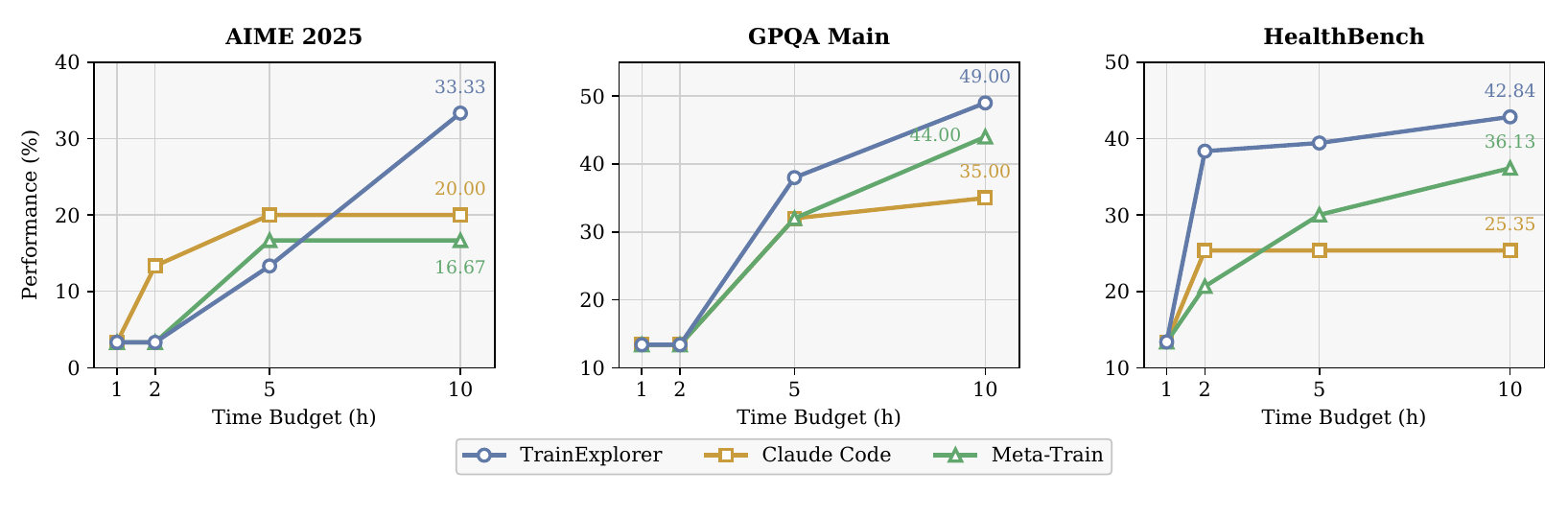}
    \vspace{-2.5em}
    \caption{Performance of TrainExplorer and baseline agents on AIME 2025, GPQA Main, and HealthBench under varying time budgets. Curves show the best evaluation scores achieved within each time budget.}
    \label{fig:time-budget}
\end{figure*}

\begin{figure}[!ht]
    \centering
    \includegraphics[width=\linewidth]{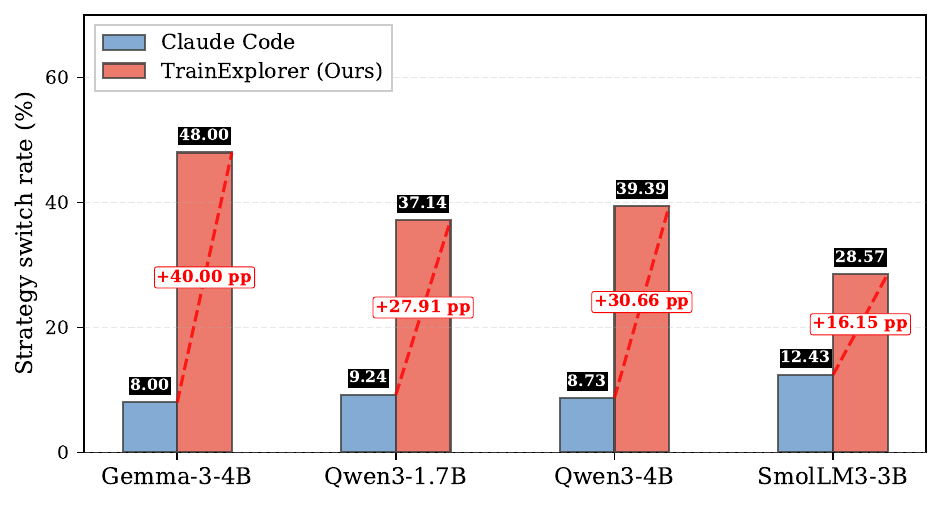}
    \vspace{-2.0em}
    \caption{Strategy switch rates of TrainExplorer and Claude Code across four base models. Red annotations indicate differences in percentage points.}
    \label{fig:strategy_switch}
\end{figure}

\paragraph{Comparison with Autonomous Post-Training Agents.}
Across all four base models, TrainExplorer achieves the highest aggregate score among autonomous post-training agents.
Notably, TrainExplorer exceeds Meta-Train by 6.16 points in overall score with the same agent LLM.
These gains may be attributed to two aspects of our design: (1) Our strategy planning draws on scientific literature to construct a candidate strategy space beyond methods familiar to the agent. (2) Execution decoupling extends reflection through single-component interventions that link performance changes to specific modifications, guiding strategy refinement.

\begin{figure*}[!ht]
    \centering
    \includegraphics[width=\linewidth]{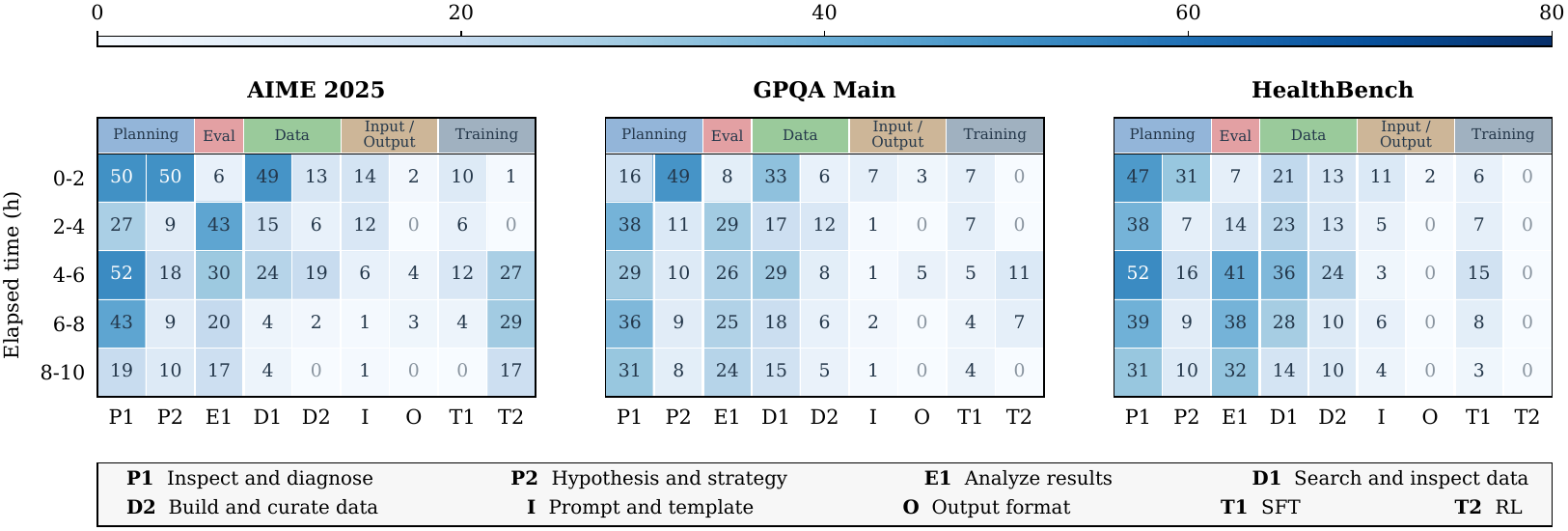}
    \caption{Tool-use counts of TrainExplorer on AIME 2025, GPQA Main, and HealthBench using Qwen3-1.7B-Base, grouped by behavior type and two-hour intervals.}
    \label{fig:agent-behavior}
\end{figure*}

\subsection{Ablation Study}
\label{sec:ablation}

To evaluate the effects of strategy planning and execution decoupling, we ablate each module in turn using Qwen3-1.7B, as shown in Table~\ref{tab:ablation-results}.

\paragraph{Effect of Strategy Planning.}
Removing strategy planning leads to a 5.26 points drop in the weighted aggregate score, with substantial performance declines across multiple benchmarks.
Notably, challenging tasks such as AIME require training strategies tailored to their reasoning demands and difficulty, and strategy planning draws on scientific literature to identify suitable candidates beyond the agent's familiar methods.
These results indicate that strategy planning helps mitigate strategy lock-in and guides exploration toward more effective training strategies.

\paragraph{Effect of Execution Decoupling.}
Removing execution decoupling reduces the weighted aggregate score by 3.55, even when experiment history and feedback are retained.
In particular, despite the availability of a candidate strategy space, the GSM8K score drops by 5.13, with coupled changes to data preparation and training obscuring the effects of individual operations.
These results indicate that execution decoupling enables effect attribution and helps distill reusable training lessons from feedback to guide strategy exploration.

\subsection{Experimental Analysis}

\paragraph{Strategy Switch.}
To examine whether TrainExplorer revises its training strategy during exploration, we calculate the strategy switch rate by measuring how often the training objective or data source changes between consecutive training runs.
Adjustments to learning rates or data formatting within the same objective and data source do not count as strategy switches.
As shown in Figure~\ref{fig:strategy_switch}, TrainExplorer exhibits a higher switch rate than Claude Code on all four base models, with overall switch rates of 37.50\% and 9.49\%, respectively, corresponding to a difference of 28.01.
These results demonstrate that TrainExplorer mitigates strategy lock-in and extends exploration beyond local adjustments within its initial training strategy.

\paragraph{Effect of Time Budget.}
We report the best-so-far online evaluation scores on AIME 2025, GPQA, and HealthBench as the time budget increases.
As shown in Figure~\ref{fig:time-budget}, several baselines show no further improvement after the early stages, whereas TrainExplorer continues to improve across all three tasks during the second half of the budget.
For example, on AIME 2025, Claude Code and Meta-Train show no further gains after five hours, while TrainExplorer improves from 13.33\% at five hours to 33.33\% at ten hours.
These results suggest that TrainExplorer continues to explore and revise training strategies in the later stages, enabling further performance gains with additional time.

\paragraph{Agent Behavior Analysis.}
To examine whether strategy exploration persists into the later stages of a run, we analyze TrainExplorer's tool calls across three tasks with Qwen3-1.7B-Base.
We classify the calls by their primary activity: problem diagnosis, strategy design, result analysis, data processing, and training-related activities.
As shown in Figure~\ref{fig:agent-behavior}, strategy design is most frequent early in the run but continues into later stages, while problem diagnosis and result analysis recur throughout the process.
These activities suggest that TrainExplorer continues to analyze experimental outcomes and consider alternative training strategies after the initial plan is established, providing a basis for subsequent revisions.
This pattern is consistent with the performance gains observed in the second half of the time-budget experiments.

\begin{figure}[!ht]
    \centering
    \includegraphics[width=\linewidth]{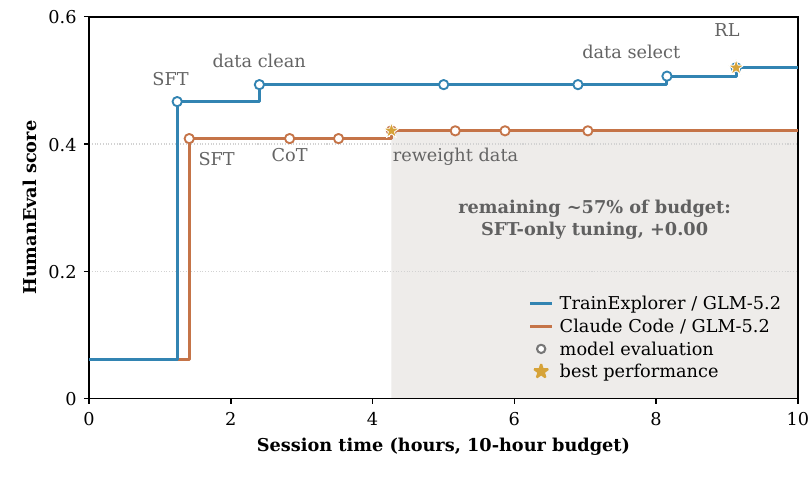}
        \vspace{-2.0em}
    \caption{Search trajectories of TrainExplorer and Claude Code on HumanEval. The curves show best-so-far evaluation scores. The shaded region highlights Claude Code's continued use of SFT alone after reaching its best score, without further improvement.}
    \label{fig:case-study}
\end{figure}

\paragraph{Case Study.}
We compare the search trajectories of TrainExplorer and Claude Code on HumanEval in Figure~\ref{fig:case-study}.
Claude Code reaches its best score at approximately 4.27 hours, after which adjustments to the SFT data mixture yield no further gains.
TrainExplorer explores both SFT and RL strategies and revises its search direction using experimental feedback.
Executioin Decoupling enables comparisons of alternative data strategies while holding the SFT component fixed, helping identify the benefits and limitations of each intervention.
This iterative process shows how TrainExplorer uses strategy planning to explore alternatives beyond its initial choice and execution decoupling to guide subsequent training strategy revision.

\section{Conclusion}
\label{sec:conclu}

In this work, we introduce TrainExplorer, a multi-agent framework for autonomous LLM post-training that integrates strategy planning and execution decoupling.
Strategy planning constructs a task-specific candidate strategy space from scientific literature for broader strategy exploration.
Execution decoupling enables single-component interventions and effect attribution, turning experimental feedback into reusable lessons for subsequent exploration.
Experiments on PostTrainBench demonstrate that TrainExplorer substantially outperforms both general-purpose agents and autonomous post-training agents.

\section*{Limitations}
Although TrainExplorer improves autonomous LLM post-training through strategy planning and execution decoupling, it has several limitations.

First, our main evaluation follows the PostTrainBench protocol, covering seven tasks and four base models with 1.7B--4B parameters under a ten-hour budget on a single H100 GPU. While this setting enables comparisons under a standardized resource budget, it focuses on optimizing individual tasks with relatively small models. Generalization to larger models, distributed training, and general-purpose post-training remains to be investigated.

Second, strategy planning depends on the availability of relevant literature in the training experience library. In emerging or specialized domains with limited reference material, the library may provide insufficient guidance for constructing candidate strategies. Extending the library's coverage remains a direction for future work.

\section*{Ethical Considerations}
TrainExplorer aims to reduce the human effort required for LLM post-training and support broader exploration of training strategies. By organizing literature knowledge and experimental feedback, the framework can help researchers investigate and adapt training methods with less manual intervention. We acknowledge the dual-use implications of this capability: simplifying model adaptation can support beneficial applications while also facilitating models optimized for harmful purposes. We encourage responsible use of TrainExplorer and human review of trained models before deployment in sensitive applications.

\section*{Use of AI Assistants}
AI assistants are utilized only for grammatical checking and language refinement, in strict adherence to the ACL policy on AI writing assistance. All ideas and technical content in this paper remain the original work of the authors.

\bibliography{custom}

\clearpage
\appendix

\begin{center}
{\Large \textbf{Appendix}}
\end{center}

\setcounter{section}{0}
\renewcommand{\thesection}{\Alph{section}}
\tableofcontents

\clearpage

\section{Experimental Details}
\label{sec:appendix-experimental-details}

\subsection{Configuration.}
We develop TrainExplorer on top of the Claude Code framework and use GLM-5.2 for the Planning, Idea, Builder, and Critic agents.
The entire run is subject to a 10-hour wall-clock budget, including strategy planning, data processing, model training, and online evaluation.
Strategy planning is performed once at the beginning of each run.

\subsection{Baselines.}
We compare against two types of baselines: general-purpose agents and autonomous post-training agents.
Results for general-purpose agents are taken from the publicly released PostTrainBench results~\cite{rank2026posttrainbench}.
We consider three autonomous post-training baselines: Meta-Train, ANDES, and AutoTrainess.

\noindent \textbf{Meta-Train.}
We adapt Meta-Harness~\cite{lee2026metaharness} to autonomous post-training, where the agent repeatedly proposes candidate strategies, trains and evaluates models, and logs the results.
The agent selectively inspects prior candidate implementations, execution traces, and evaluation scores to diagnose failures and propose or revise post-training pipelines.

\noindent \textbf{ANDES.}
ANDES~\cite{zhao2026andes} exposes data synthesis as an interactive tool for trainer agents, using a self-evolving world tree to generate diverse training data aligned with target capabilities.
The trainer agent uses diagnostic reports returned by the tool to adjust subsequent synthesis configurations, forming a closed-loop data synthesis process.

\noindent \textbf{AutoTrainess.}
AutoTrainess~\cite{yu2026autotrainess} externalizes prior human experience as explicit workflows, rules, and execution constraints within structured interfaces for autonomous training.
These interfaces guide the agent through planning, data preparation, training, and evaluation, with experimental feedback informing subsequent refinement of the post-training pipeline.

\subsection{Evaluation.}
We select the model with the highest evaluation score during iterative search as the final model.
To reduce evaluation noise, we evaluate the selected model three times on the full evaluation set and report the mean score.

Let $r_{m,j}$ denote the final evaluation score for base model $m$ on task $j$, with $N_{\mathrm{model}}=4$ base models and $N_{\mathrm{task}}=7$ tasks.
Following PostTrainBench~\cite{rank2026posttrainbench}, we aggregate scores in two stages.
First, we average the scores for each task across the four base models.
We then compute a weighted average across tasks:
\begin{equation}
    \bar{r}_j = \frac{1}{N_{\mathrm{model}}}
    \sum_{m=1}^{N_{\mathrm{model}}} r_{m,j}.
\label{eq:appendix-per-task-score}
\end{equation}
\begin{equation}
    \mathrm{AVG} = \sum_{j=1}^{N_{\mathrm{task}}} w_j \bar{r}_j.
\label{eq:appendix-aggregate-score}
\end{equation}
We use the official PostTrainBench task weights, obtained by normalizing the inverse score gaps between the official instruction-tuned and base models:
\begin{equation}
    w_j =
    \frac{
        \bigl(\bar{r}_j^{\mathrm{instruct}}
        - \bar{r}_j^{\mathrm{base}}\bigr)^{-1}
    }{
        \displaystyle\sum_{\ell=1}^{N_{\mathrm{task}}}
        \bigl(\bar{r}_{\ell}^{\mathrm{instruct}}
        - \bar{r}_{\ell}^{\mathrm{base}}\bigr)^{-1}
    }.
\label{eq:appendix-task-weights}
\end{equation}
Here, $\bar{r}_j^{\mathrm{instruct}}$ and $\bar{r}_j^{\mathrm{base}}$ denote the scores of the official instruction-tuned and base models on task $j$, respectively, averaged across the four model configurations.
These weights assign greater weight to tasks on which instruction tuning yields smaller gains.
For results taken from prior papers, we retain the task scores and aggregate scores as originally reported.

\begin{table*}[t]
\tablestyle{5pt}{1.3}
\resizebox{\linewidth}{!}{%
\begin{tabular}{cc|ccccccc|c}
\toprule
\textbf{Method} & \textbf{Model} & AIME & ArenaHard & BFCL & GPQA & GSM8K & HealthBench & HumanEval & AVG \\
\midrule \midrule
\multicolumn{2}{c|}{Base Models} & 0.00 & 0.91 & 0.00 & 14.06 & 12.66 & 7.54 & 7.93 & 6.66 \\
\multicolumn{2}{c|}{Official Instruct Models} & 26.67 & 50.00 & 94.00 & 35.49 & 88.48 & 44.92 & 68.90 & 49.41 \\
\midrule
\rowcolor{COLOR_MEAN}
\multicolumn{10}{l}{\textit{General CLI Agent}} \\
Codex & GPT-5.6 Sol & 5.00 & 27.47 & 94.50 & 29.13 & 59.82 & 26.89 & 66.77 & 34.85 \\
Claude Code & Opus 4.7 & 6.67 & 31.26 & 91.33 & 29.32 & 50.47 & 17.03 & 42.28 & 30.08 \\
Claude Code & Kimi K3 & 4.44 & 32.37 & 0.00 & 27.83 & 77.23 & 32.47 & 62.60 & 30.03 \\
Claude Code & Opus 4.8 & 3.33 & 11.11 & 91.00 & 25.89 & 64.40 & 25.27 & 39.02 & 29.19 \\
Codex & GPT-5.5 & 3.33 & 9.18 & 46.00 & 28.01 & 51.18 & 21.02 & 50.00 & 25.28 \\
Cursor & Grok 4.5 & 5.00 & 0.91 & 0.00 & 28.68 & 70.09 & 9.08 & 50.61 & 21.26 \\
OpenCode & Gemini 3.1 Pro & 0.00 & 2.27 & 85.67 & 16.15 & 39.58 & 9.26 & 36.18 & 19.47 \\
Codex & GPT-5.4 & 0.00 & 0.47 & 0.00 & 29.39 & 50.42 & 15.71 & 30.28 & 17.47 \\
\midrule
\rowcolor{COLOR_MEAN}
\multicolumn{10}{l}{\textit{Autonomous PostTraining Agent}} \\
ANDES & GLM-4.7 & 0.00 & 12.94 & 89.00 & 27.03 & 61.91 & 25.87 & 56.14 & 30.39 \\
AutoTrainess & GPT-5.4 & 0.00 & 0.32 & 90.00 & 29.91 & 54.74 & 14.95 & 40.85 & 25.67 \\
\midrule
Claude Code & GLM-5.2 & 1.11 & 13.51 & \textbf{94.00} & 29.32 & 79.61 & 25.22 & 55.28 & 33.03 \\
Meta-Train & GLM-5.2 & 2.22 & \textbf{28.20} & 91.00 & 31.55 & 74.63 & \textbf{27.48} & 56.71 & 34.99 \\
\rowcolor{LightCyan}
\textbf{TrainExplorer} & GLM-5.2 & \textbf{6.67} & 27.66 & \textbf{94.00} & \textbf{32.07} & \textbf{81.05} & 27.36 & \textbf{65.24} & \textbf{37.77} \\
\bottomrule
\end{tabular}
}
\caption{Performance (\%) of post-training agents on PostTrainBench using Qwen3-1.7B-Base as the base model. Base and official instruct models are included as references. AVG denotes the weighted aggregate score across seven tasks. Bold indicates the best score in each column among methods using GLM-5.2 as the agent LLM.}
\label{tab:main-results-qwen3-1-7b}
\end{table*}

\section{Per-Base-Model Results}
\label{sec:main-results-by-model}
We report detailed results for Qwen3-1.7B, Qwen3-4B, SmolLM3-3B, and Gemma-3-4B in Tables~\ref{tab:main-results-qwen3-1-7b}, \ref{tab:main-results-qwen3-4b}, \ref{tab:main-results-smollm3-3b}, and \ref{tab:main-results-gemma3-4b}, respectively.
TrainExplorer outperforms all autonomous post-training baselines across all four base models.

\paragraph{Qwen3-1.7B.}
TrainExplorer achieves a weighted aggregate score of 37.77, exceeding Meta-Train, the strongest baseline, by 2.78 percentage points.
Among methods using the same agent LLM, TrainExplorer achieves higher scores on AIME, GPQA, GSM8K, and HumanEval.
On HumanEval, it reaches 65.24\%, outperforming Claude Code by 9.96 percentage points.
This approaches the official instruction-tuned model's score of 68.90\%, showing that TrainExplorer can substantially narrow the code generation performance gap between a small base model and its official instruction-tuned counterpart within the allotted budget.

\paragraph{Qwen3-4B.}
TrainExplorer achieves the highest weighted aggregate score of 49.53, outperforming Meta-Train by 6.09 percentage points and Claude Code using the same agent LLM by 10.66 percentage points.
Its AIME and HealthBench scores exceed those of Claude Code by 10.00 and 14.72 percentage points, respectively.
Together with the Qwen3-1.7B results, these gains demonstrate the effectiveness of TrainExplorer across different model sizes within the same model family.

\paragraph{SmolLM3-3B.}
TrainExplorer achieves a weighted aggregate score of 35.66, exceeding ANDES, the strongest autonomous post-training baseline, at 31.91.
Among methods using the same agent LLM, TrainExplorer achieves the highest scores on GPQA, GSM8K, and HumanEval.
These gains in scientific question answering, mathematical reasoning, and code generation extend the findings beyond the Qwen family, further supporting the applicability of TrainExplorer across base model families.

\paragraph{Gemma-3-4B.}
TrainExplorer achieves a weighted aggregate score of 33.14, slightly exceeding the strongest baseline, Claude Code (Opus 4.8), at 32.87.
Compared with Claude Code using the same agent LLM, TrainExplorer improves the weighted aggregate score by 8.16 percentage points and achieves gains on six tasks.
Its BFCL score increases from 35.00\% to 88.00\%, while ArenaHard, GSM8K, and HumanEval improve by 11.75, 8.61, and 7.52 percentage points, respectively.
These gains span function calling, writing, mathematical reasoning, and code generation, supporting TrainExplorer's ability to identify effective post-training strategies for Gemma across diverse task requirements.

\begin{table*}[t]
\tablestyle{5pt}{1.3}
\resizebox{\linewidth}{!}{%
\begin{tabular}{cc|ccccccc|c}
\toprule
\textbf{Method} & \textbf{Model} & AIME & ArenaHard & BFCL & GPQA & GSM8K & HealthBench & HumanEval & AVG \\
\midrule \midrule
\multicolumn{2}{c|}{Base Models} & 3.33 & 3.42 & 0.00 & 13.39 & 41.85 & 13.38 & 36.59 & 14.34 \\
\multicolumn{2}{c|}{Official Instruct Models} & 53.33 & 86.84 & 95.00 & 44.64 & 93.78 & 52.72 & 77.44 & 63.75 \\
\midrule
\rowcolor{COLOR_MEAN}
\multicolumn{10}{l}{\textit{General CLI Agent}} \\
Codex & GPT-5.6 Sol & 10.00 & 32.86 & 94.50 & 34.04 & 80.06 & 27.07 & 82.62 & 41.18 \\
Claude Code & Kimi K3 & 22.22 & 48.25 & 32.67 & 32.22 & 86.43 & 29.51 & 79.47 & 41.03 \\
Claude Code & Opus 4.8 & 23.33 & 48.67 & 0.00 & 31.47 & 89.88 & 34.06 & 69.21 & 38.78 \\
Cursor & Grok 4.5 & 20.00 & 29.71 & 0.00 & 25.56 & 74.91 & 19.16 & 81.10 & 32.10 \\
Claude Code & Opus 4.7 & 7.78 & 21.90 & 62.00 & 24.03 & 70.58 & 21.72 & 62.60 & 31.01 \\
Codex & GPT-5.5 & 6.67 & 9.98 & 46.00 & 34.04 & 60.77 & 15.63 & 71.04 & 29.60 \\
OpenCode & Gemini 3.1 Pro & 2.22 & 4.98 & 57.33 & 20.16 & 47.54 & 16.46 & 52.03 & 22.76 \\
Codex & GPT-5.4 & 2.22 & 6.53 & 31.00 & 23.96 & 51.58 & 16.57 & 25.81 & 19.41 \\
\midrule
\rowcolor{COLOR_MEAN}
\multicolumn{10}{l}{\textit{Autonomous PostTraining Agent}} \\
ANDES & GLM-4.7 & 9.00 & 50.43 & 90.00 & 30.21 & 82.17 & 29.69 & 65.35 & 40.18 \\
AutoTrainess & GPT-5.4 & 6.67 & 1.41 & 100.00 & 35.71 & 75.59 & 18.20 & 47.56 & 32.60 \\
\midrule
Claude Code & GLM-5.2 & 16.67 & 54.25 & 31.67 & 32.89 & 85.29 & 23.86 & 76.02 & 38.87 \\
Meta-Train & GLM-5.2 & 14.44 & 53.65 & \textbf{94.00} & 32.96 & \textbf{90.32} & 27.22 & 70.12 & 43.44 \\
\rowcolor{LightCyan}
\textbf{TrainExplorer} & GLM-5.2 & \textbf{26.67} & \textbf{56.11} & \textbf{94.00} & \textbf{34.30} & 88.80 & \textbf{38.58} & \textbf{78.05} & \textbf{49.53} \\
\bottomrule
\end{tabular}
}
\caption{Performance (\%) of post-training agents on PostTrainBench using Qwen3-4B-Base as the base model. Base and official instruct models are included as references. AVG denotes the weighted aggregate score across seven tasks. Bold indicates the best score in each column among methods using GLM-5.2 as the agent LLM.}
\label{tab:main-results-qwen3-4b}
\end{table*}

\begin{table*}[t]
\tablestyle{5pt}{1.3}
\resizebox{\linewidth}{!}{%
\begin{tabular}{cc|ccccccc|c}
\toprule
\textbf{Method} & \textbf{Model} & AIME & ArenaHard & BFCL & GPQA & GSM8K & HealthBench & HumanEval & AVG \\
\midrule \midrule
\multicolumn{2}{c|}{Base Models} & 3.33 & 0.42 & 0.00 & 4.91 & 21.08 & 0.00 & 6.10 & 4.52 \\
\multicolumn{2}{c|}{Official Instruct Models} & 26.67 & 49.20 & 84.00 & 33.26 & 82.18 & 29.58 & 70.12 & 44.81 \\
\midrule
\rowcolor{COLOR_MEAN}
\multicolumn{10}{l}{\textit{General CLI Agent}} \\
Codex & GPT-5.6 Sol & 15.00 & 13.57 & 93.50 & 30.47 & 74.98 & 27.51 & 56.10 & 36.48 \\
Claude Code & Opus 4.8 & 16.67 & 37.19 & 47.00 & 27.68 & 51.52 & 32.63 & 28.96 & 30.77 \\
Claude Code & Kimi K3 & 18.89 & 28.49 & 0.00 & 16.07 & 76.40 & 34.92 & 58.94 & 30.30 \\
Codex & GPT-5.5 & 5.00 & 7.95 & 46.00 & 33.15 & 64.03 & 20.87 & 46.34 & 27.48 \\
Claude Code & Opus 4.7 & 10.00 & 10.21 & 62.33 & 27.83 & 66.11 & 14.32 & 35.57 & 26.69 \\
OpenCode & Gemini 3.1 Pro & 12.22 & 6.42 & 27.67 & 18.01 & 55.42 & 18.45 & 38.01 & 22.08 \\
Cursor & Grok 4.5 & 16.67 & 7.78 & 0.00 & 27.57 & 59.78 & 0.00 & 34.45 & 19.92 \\
Codex & GPT-5.4 & 0.00 & 16.77 & 29.67 & 29.02 & 50.14 & 6.11 & 24.19 & 18.63 \\
\midrule
\rowcolor{COLOR_MEAN}
\multicolumn{10}{l}{\textit{Autonomous PostTraining Agent}} \\
ANDES & GLM-4.7 & 6.60 & 14.11 & 78.67 & 25.80 & 67.52 & 37.20 & 41.37 & 31.91 \\
AutoTrainess & GPT-5.4 & 0.00 & 0.10 & 97.00 & 28.13 & 57.70 & 18.82 & 29.88 & 25.60 \\
\midrule
Claude Code & GLM-5.2 & \textbf{13.33} & 13.52 & 31.00 & 29.76 & 64.06 & \textbf{31.51} & 45.93 & 29.91 \\
Meta-Train & GLM-5.2 & 0.00 & \textbf{19.83} & \textbf{92.00} & 28.20 & 64.70 & 22.46 & 45.73 & 30.03 \\
\rowcolor{LightCyan}
\textbf{TrainExplorer} & GLM-5.2 & 12.22 & 16.77 & \textbf{92.00} & \textbf{30.43} & \textbf{73.59} & 28.24 & \textbf{52.64} & \textbf{35.66} \\
\bottomrule
\end{tabular}
}
\caption{Performance (\%) of post-training agents on PostTrainBench using SmolLM3-3B-Base as the base model. Base and official instruct models are included as references. AVG denotes the weighted aggregate score across seven tasks. Bold indicates the best score in each column among methods using GLM-5.2 as the agent LLM.}
\label{tab:main-results-smollm3-3b}
\end{table*}

\begin{table*}[t]
\tablestyle{5pt}{1.3}
\resizebox{\linewidth}{!}{%
\begin{tabular}{cc|ccccccc|c}
\toprule
\textbf{Method} & \textbf{Model} & AIME & ArenaHard & BFCL & GPQA & GSM8K & HealthBench & HumanEval & AVG \\
\midrule \midrule
\multicolumn{2}{c|}{Base Models} & 0.00 & 0.29 & 6.00 & 1.56 & 6.14 & 17.04 & 0.61 & 4.60 \\
\multicolumn{2}{c|}{Official Instruct Models} & 10.00 & 94.80 & 67.00 & 31.47 & 83.55 & 46.06 & 69.51 & 46.58 \\
\midrule
\rowcolor{COLOR_MEAN}
\multicolumn{10}{l}{\textit{General CLI Agent}} \\
Claude Code & Opus 4.8 & 0.00 & 47.39 & 50.50 & 28.35 & 69.67 & 34.77 & 52.13 & 32.87 \\
Codex & GPT-5.6 Sol & 0.00 & 31.68 & 95.00 & 26.90 & 63.04 & 28.08 & 50.30 & 32.40 \\
Codex & GPT-5.5 & 0.00 & 24.45 & 86.50 & 27.46 & 50.83 & 21.62 & 28.05 & 26.55 \\
Claude Code & Kimi K3 & 1.11 & 37.72 & 6.00 & 24.78 & 65.20 & 30.44 & 48.17 & 26.49 \\
Claude Code & Opus 4.7 & 1.11 & 14.99 & 92.00 & 28.72 & 55.70 & 18.16 & 27.85 & 26.43 \\
OpenCode & Gemini 3.1 Pro & 1.11 & 16.01 & 83.67 & 19.79 & 39.50 & 21.14 & 34.55 & 23.64 \\
Cursor & Grok 4.5 & 0.00 & 18.10 & 6.00 & 26.45 & 55.69 & 17.04 & 39.02 & 20.52 \\
Codex & GPT-5.4 & 0.00 & 14.52 & 33.67 & 29.54 & 40.59 & 17.08 & 29.07 & 20.49 \\
\midrule
\rowcolor{COLOR_MEAN}
\multicolumn{10}{l}{\textit{Autonomous PostTraining Agent}} \\
ANDES & GLM-4.7 & 0.00 & 25.16 & 85.00 & 25.42 & 76.57 & 33.28 & 32.11 & 31.02 \\
AutoTrainess & GPT-5.4 & 0.00 & 16.76 & 100.00 & 22.99 & 40.64 & 12.22 & 34.76 & 23.88 \\
\midrule
Claude Code & GLM-5.2 & \textbf{0.00} & 26.65 & 35.00 & 26.64 & 61.87 & 25.59 & 32.72 & 24.98 \\
Meta-Train & GLM-5.2 & \textbf{0.00} & 31.57 & 6.00 & 25.82 & 63.84 & 22.90 & 34.76 & 22.98 \\
\rowcolor{LightCyan}
\textbf{TrainExplorer} & GLM-5.2 & \textbf{0.00} & \textbf{38.40} & \textbf{88.00} & \textbf{30.36} & \textbf{70.48} & \textbf{29.40} & \textbf{40.24} & \textbf{33.14} \\
\bottomrule
\end{tabular}
}
\caption{Performance (\%) of post-training agents on PostTrainBench using Gemma-3-4B-PT as the base model. Base and official instruct models are included as references. AVG denotes the weighted aggregate score across seven tasks. Bold indicates the best score in each column among methods using GLM-5.2 as the agent LLM.}
\label{tab:main-results-gemma3-4b}
\end{table*}

\end{document}